\documentclass[letterpaper, 10 pt, conference]{ieeeconf}  % Comment this line out if you need a4paper
\IEEEoverridecommandlockouts                              % This command is only needed if 
\usepackage{graphicx}
\usepackage{amsmath} % assumes amsmath package installed
\usepackage{amssymb}  % assumes amsmath package installed
\usepackage{diagbox}
\usepackage{booktabs}
\usepackage{multirow}
\usepackage{float}
\usepackage{caption}
\usepackage{capt-of}
\usepackage{colortbl}
\usepackage{todonotes}
\usepackage{hyperref}
\usepackage{stfloats}
\usepackage{xcolor}
\hypersetup{pdfborder={0 0 0}}

\title{\LARGE \bf
WB-WAM: Heterogeneous Body-Hand Pre-training \\ for Humanoid Loco-Manipulation
}

\author{%
Chuan Qin$^{1,2,3,*}$ \quad
Shaoting Zhu$^{1,2,*}$ \quad
Siyuan Luo$^{2,3}$ \quad
Siqiao Huang$^{1}$ \quad \\
Hongyu Zhao$^{2}$ \quad
Shanaka Baduge $^{3}$ \quad
Hang Zhao$^{1,2,\dagger}$\\[2pt]
Project Page: \mbox{\href{https://wb-wam.github.io/}{\texttt{WB-WAM.github.io}}}%
\thanks{%
$^{1}$IIIS, Tsinghua University, Beijing, China.}%
\thanks{%
$^{2}$Xiong'an Institute of Artificial Intelligence, Xiong'an, China.}%
\thanks{%
$^{3}$The University of Melbourne, Melbourne, Australia.}%
\thanks{%
$^{*}$These authors contributed equally to this work.}%
\thanks{%
$^{\dagger}$Corresponding author.
E-mail: \texttt{hangzhao@mail.tsinghua.edu.cn}}%
}

\newcommand{\inserttitleteaser}{%
  \par\noindent
  \begin{minipage}{\textwidth}
    \centering
    \vspace{-4mm}
    \includegraphics[width=0.97\linewidth]{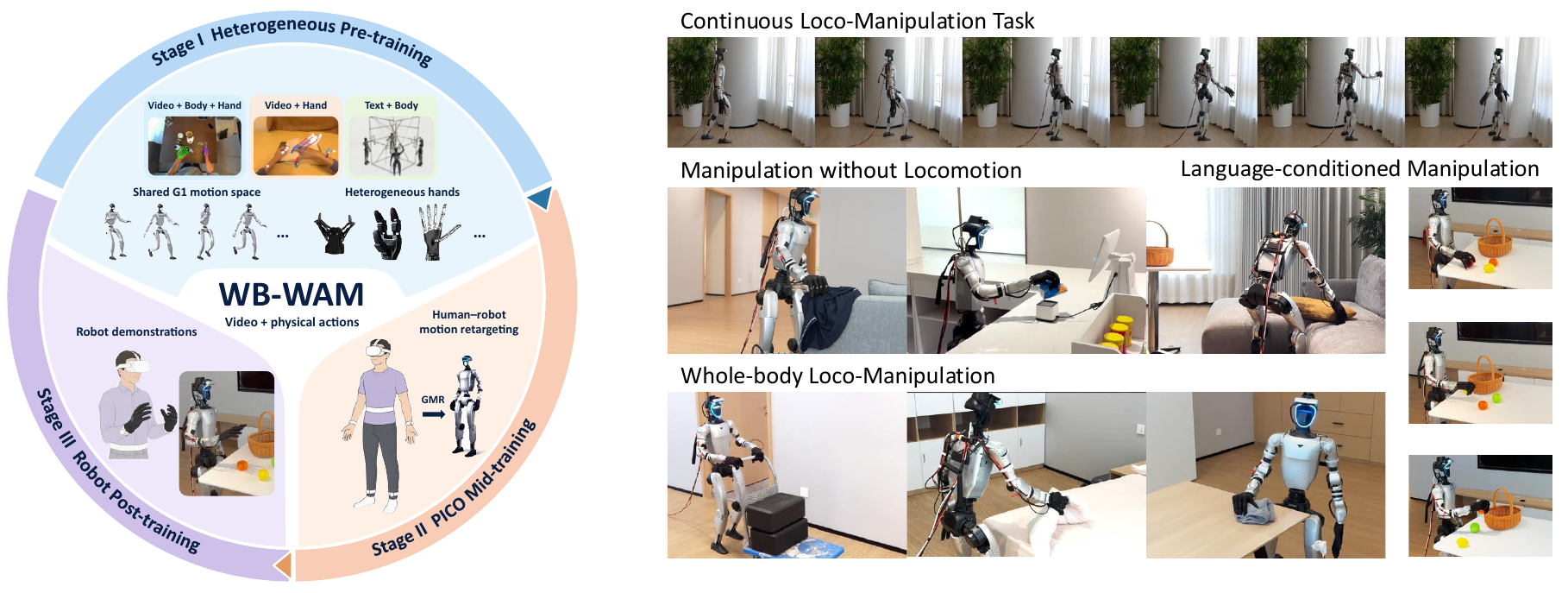}
    \captionof{figure}{\textbf{Overview of WB-WAM.}
    Left: Heterogeneous video and motion pre-training, PICO mid-training,
    and task-specific robot adaptation share an explicit physical action
    representation for body and dexterous hand prediction.
    Right: Real-world deployments cover locomotion and manipulation, together with language-conditioned fruit selection
and placement.}
    \label{fig:teaser}
  \end{minipage}
  \par\vspace{0.5\baselineskip}
}

\makeatletter
\@ifundefined{IEEEaftertitletext}{%
  \g@addto@macro\@maketitle{\inserttitleteaser}%
}{%
  \IEEEoverridecommandlockouts
  \IEEEaftertitletext{\inserttitleteaser}%
}
\makeatother

\begin{document}

\maketitle
\thispagestyle{empty}
\pagestyle{empty}

\begin{abstract}
Humanoid loco-manipulation demands coordinated body and hand behavior, while conventional robot pre-training data provide limited coverage of such whole-body motion. We present WB-WAM, a World Action Model that incorporates explicit whole-body action supervision into generative video pre-training. A shared physical action space integrates body, root, and dexterous hand annotations from heterogeneous sources, enabling joint video and action learning from 1880.2 hours of partially annotated video and motion data. The resulting priors are refined through PICO mid-training and adapted to robot tasks with auxiliary forward kinematics supervision. We construct WB-Datasets to support these stages with retargeted egocentric human demonstrations and robot trajectories, allowing task-aligned human motion to supplement limited robot data. Evaluations in simulation demonstrate strong whole-body task performance with 81.9\% in HumanoidArena, while real-world experiments further validate WB-WAM with 84.0\% mean success across five tasks. Moreover, task-aligned PICO mid-training improves downstream task performance while reducing the need for real-robot demonstrations. These results support heterogeneous whole-body pre-training and human motion transfer as a practical route to data-efficient humanoid loco-manipulation.
\end{abstract}

\section{INTRODUCTION}

General-purpose humanoid robots must coordinate body motion and dexterous manipulation in response to visual observations and task instructions. Large-scale robot pre-training has established reusable manipulation priors, with influential efforts drawing primarily on single-arm, bimanual, and wheeled mobile systems~\cite{o2024open,black2024pi_0,qin2026comprehensive}. Their data and action interfaces provide limited direct supervision of coordinated humanoid body, root, and hand motion. Extending these priors to humanoid loco-manipulation therefore requires attention to the coverage of whole-body actions during pre-training, beyond the diversity of tasks and objects represented in the data.

Recent advances in humanoid locomotion have pushed toward generalist control across diverse motion skills~\cite{huang2026omg,zhu2026hiking}. Meanwhile, recent humanoid foundation models learn from heterogeneous human and robot demonstrations~\cite{bjorck2025gr00t,wei2026psi_0,jiang2026wholebodyvla}. In parallel, World Action Models (WAMs) couple action generation with predictive visual modeling, with recent systems extending this formulation to humanoid control~\cite{yuan2026fast,zheng2026motionwam,li2026omega0}. These advances motivate pre-training that jointly models visual dynamics and explicit whole-body actions. A central challenge is to exploit complementary motion sources without requiring every training sequence to contain complete video, body, and hand annotations.

Egocentric manipulation recordings may provide detailed hand annotations without body motion, while motion collections may contain whole-body trajectories without accompanying video. Combining these sources calls for a prediction interface that accommodates their complementary supervision. We propose WB-WAM, a WAM that incorporates explicit whole-body action supervision into generative video pre-training. Building on Fast-WAM~\cite{yuan2026fast}, the model jointly learns visual dynamics and action trajectories, representing actions in a shared 72-D physical space comprising body joint references, root motion, and articulated hand states. Available annotations from separate sources populate their corresponding coordinates, allowing pre-training to use data without complete video and action labels.

Whole-body pre-training is followed by motion transfer and robot adaptation. Stage I learns from approximately 1,900 hours of heterogeneous video and motion data. Stage II uses retargeted PICO demonstrations to specialize these priors through egocentric observations paired with our shared action space. Task-aligned demonstrations in the transfer study supply interaction experience before robot adaptation, allowing human motion supervision to complement a smaller set of robot demonstrations. WB-Datasets supports this transfer with 22 hours of PICO demonstrations and 1,011 robot episodes across eight tasks.

Stage III adapts the pre-trained model to downstream robot tasks using task-specific demonstrations and auxiliary forward kinematics supervision. At deployment, predicted body and root references are executed through SONIC~\cite{luo2026sonic}, while hand references remain direct joint commands.

In simulation, WB-WAM achieves a SOTA mean
success rate of 81.9\% on HumanoidArena, surpassing the task-wise best reported baselines on all seven tasks. On five real-robot tasks, WB-WAM without PICO mid-training achieves 84.0\% mean success, compared with 80.0\% for the strongest evaluated baseline, OpenWAM. On four tasks with aligned PICO demonstrations, mid-training followed by adaptation with 30 robot demonstrations per task achieves 73.8\% mean success, exceeding direct adaptation with 100 demonstrations at 65.0\%. This comparison uses 70\% fewer robot demonstrations, supplemented by human demonstrations of the same tasks. A shared fruit-manipulation policy further demonstrates target selection from language instructions. 

The main contributions are:
\begin{itemize}
    \item A whole-body humanoid pre-training framework that couples
    generative video modeling with explicit body, root, and dexterous hand
    supervision at scale, using a shared physical action space to integrate
    heterogeneous motion annotations.

    \item A three-stage training recipe and WB-Datasets that use retargeted
    PICO demonstrations as intermediate supervision, linking whole-body
    pre-training to robot adaptation with fewer robot demonstrations.

    \item Extensive evaluations on HumanoidArena and real-world humanoid tasks, including comparisons against six baselines on the physical robot. Additional studies demonstrate language-conditioned manipulation and visuomotor generalization to unseen visual conditions.
\end{itemize}

\section{Related Work}

\subsection{Humanoid Foundation Models for Loco-manipulation}

Recent humanoid foundation models learn reusable visuomotor priors from heterogeneous data to reduce embodiment-specific robot data requirements~\cite{bjorck2025gr00t,wei2026psi_0,jiang2026wholebodyvla,hu2026openhlm,shi2026egohumanoid}. These models differ in how their action representations are grounded to humanoid control. GR00T N1~\cite{bjorck2025gr00t} couples a vision-language backbone with a flow-matching action policy trained on robot, simulation, and video data. $\Psi_0$~\cite{wei2026psi_0} pre-trains on tokenized bimanual task-space actions, then learns a continuous action expert from humanoid trajectories with the backbone frozen. WholeBodyVLA~\cite{jiang2026wholebodyvla} learns discrete latent actions from visual transitions without action annotations and grounds them into upper-body joint targets and locomotion commands. OpenHLM~\cite{hu2026openhlm} adapts a VLA pre-trained for nonhumanoid manipulation to whole-body reference control. Across these systems, scalable video data lack recorded actions or use proxy action representations, while controller-compatible humanoid references enter mainly through robot data or later adaptation. Directly incorporating temporally dense, physically interpretable body, root, and articulated-hand references into the large-scale pre-training of a world action model remains comparatively underexplored.

\subsection{World Action Models}

World Action Models (WAMs) couple action generation with predictive visual modeling~\cite{yuan2026fast}, building on large video and world models~\cite{agarwal2025cosmos,wan2025wan}. Approaches include inverse dynamics decoding~\cite{hu2024video,pai2025mimicvideo}, autoregressive or shared video--action representations~\cite{cen2025worldvla,zhu2025unified,bi2026motus,li2026causal}, and coupled video-action denoising~\cite{ye2026world,ma2026dit4dit,zhou2026tau_0,wang2026openwam}. Cosmos Policy adapts a pre-trained video model into a policy~\cite{kim2026cosmospolicy}, while Fast-WAM enables action inference without future video synthesis~\cite{yuan2026fast}. Egocentric human video further supports cross-embodiment co-training~\cite{li2026egowam}. Most evaluations focus on fixed-base or upper-body manipulation. Recent humanoid WAMs typically use compact kinematic plans or controller-oriented latents for loco-manipulation~\cite{zheng2026motionwam,beingbeyond2026beingm07,li2026omega0}. In contrast, WB-WAM introduces large-scale, direct whole-body action supervision during generative video pre-training. Its coordinatewise physical action space unifies body and hand annotations from separate sources to explicitly predict whole-body motion references.

\section{Methods}

\begin{figure*}[!t]
    \centering
    \vspace{1mm}
    \includegraphics[width=0.95\textwidth]{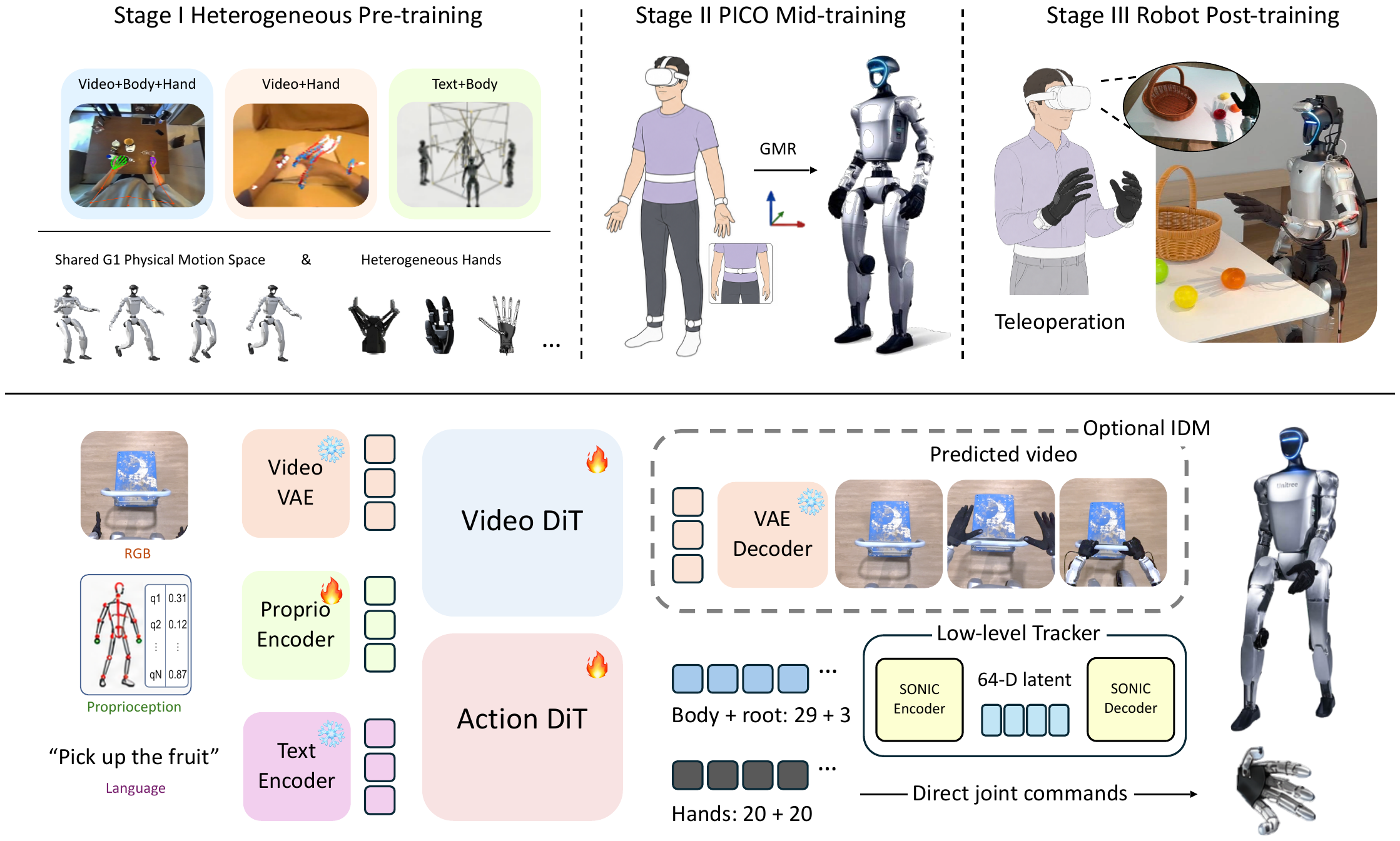}
    \caption{\textbf{WB-WAM architecture and three-stage training framework.}
    Top: Progressive training integrates heterogeneous supervision,
    retargeted PICO motion, and real-robot demonstrations.
    Bottom: Video and action experts jointly model visual dynamics and
    whole-body actions conditioned on vision, language, and proprioception.
    Body and root references are executed through SONIC, while hand
    references directly control the finger joints.}
    \label{fig:pipeline}
    \vspace{-5mm}
\end{figure*}

We propose WB-WAM, a World Action Model for humanoid loco-manipulation that incorporates explicit, physically interpretable whole-body action supervision into generative video pre-training. Its three-stage curriculum integrates heterogeneous video and action pre-training, motion transfer from egocentric human demonstrations, and adaptation to individual robot tasks (\autoref{fig:pipeline}). Stage I learns from heterogeneous data with partially observed body and hand annotations. Stage II specializes the pre-trained model using PICO demonstrations with retargeted body and hand references. Stage III adapts an independent model to each robot task and supplements action learning with explicit supervision of body geometry. The prediction space remains unchanged throughout training; the frozen SONIC encoder~\cite{luo2026sonic} is invoked only at deployment to map predicted body and root trajectories into controller latents.

\subsection{Preliminary: World Action Modeling with MoT}

WB-WAM builds on the multimodal flow matching formulation of Fast-WAM~\cite{yuan2026fast}. Its Mixture-of-Transformers (MoT) backbone comprises video and action experts with modality-specific parameters. Through MoT attention, the action expert conditions on features from the video expert. Future RGB frames are encoded by the frozen VAE of the pre-trained video model. The video expert models a flow field over these latent representations, whereas the action expert operates on normalized physical reference trajectories.

Let $\mathbf{y}_0 \in \{\mathbf{Z}_0,\mathbf{A}_0\}$ denote a clean target, with $\mathbf{Z}_0$ and $\mathbf{A}_0$ representing the encoded future video and the normalized action trajectory, respectively. For a flow time $\tau \in [0,1]$ and Gaussian noise $\boldsymbol{\epsilon} \sim \mathcal{N}(\mathbf{0},\mathbf{I})$ of matching dimensionality, the noisy sample and target flow velocity are

\begin{equation}
    \label{eq:flow_path}
    \mathbf{y}_{\tau}
    = (1-\tau)\mathbf{y}_0 + \tau\boldsymbol{\epsilon},
    \qquad
    \mathbf{u} = \boldsymbol{\epsilon} - \mathbf{y}_0.
\end{equation}

Video and action flow times, $\tau_v$ and $\tau_a$, are sampled independently.

\subsection{Problem Formulation}

For humanoid loco-manipulation, WB-WAM jointly models a future video clip $\mathbf{V}_t$ and its corresponding trajectory of whole-body physical references $\mathbf{A}_t$. At time $t$, the predictions are conditioned on an egocentric RGB observation $\mathbf{I}_t$, a proprioceptive state $\mathbf{s}_t$, and a language instruction $\ell$:

\begin{equation}
    \label{eq:conditional_joint}
    p_{\boldsymbol{\theta}}
    \left(\mathbf{V}_t,\mathbf{A}_t
    \mid \mathbf{I}_t,\mathbf{s}_t,\ell\right),
\end{equation}

where $\boldsymbol{\theta}$ denotes the trainable model parameters. Each action reference is represented as

\begin{equation}
    \label{eq:physical_reference}
    \mathbf{a}_t =
    \left[\mathbf{q}_t^B,\mathbf{r}_t,
    \mathbf{q}_t^L,\mathbf{q}_t^R\right]
    \in \mathbb{R}^{72}.
\end{equation}

Here, $\mathbf{q}_t^B \in \mathbb{R}^{29}$ contains the G1 body joint position references; $\mathbf{r}_t = (\phi_t,\theta_t,\omega_t^z) \in \mathbb{R}^3$ specifies root roll, root pitch, and yaw angular velocity; and $\mathbf{q}_t^L,\mathbf{q}_t^R \in \mathbb{R}^{20}$ contain the articulated hand references. The 72-D action vector defines a shared prediction space across all training stages.

\subsection{Stage I: Heterogeneous Body--Hand Pre-training}

We couple generative video pre-training with explicit humanoid action supervision on $\mathcal{D}_{\mathrm I}$, a heterogeneous dataset of 1880.2 hours of video and motion data. The video and action experts are jointly optimized under
this heterogeneous supervision. The dataset comprises three supervision types: video paired with body and hand motion ($\mathcal{D}_{\mathrm{VBH}}$), egocentric video paired with hand motion ($\mathcal{D}_{\mathrm{VH}}$), and text-conditioned body motion without video ($\mathcal{D}_{\mathrm{TB}}$). Body annotations include joint references and available root references. Available annotations are mapped to their corresponding channels in the physical action space of \autoref{eq:physical_reference}, allowing complementary body and hand supervision from separately sourced datasets. Data sources and curation are described in \autoref{sec:external_pre-training_data}.

A binary mask $\mathbf{M}$ identifies annotated, non-padded action entries. We apply this mask to the noisy action input in \autoref{eq:flow_path} and construct the target flow velocity as

\begin{equation}
    \label{eq:action_targets}
    \mathbf{u}_a
    = \mathbf{M}\odot
    \bigl(\boldsymbol{\epsilon}_a-\mathbf{A}_0\bigr).
\end{equation}

Here, $\odot$ denotes element-wise multiplication. The action expert predicts $\widehat{\mathbf{u}}_a$ from the noisy action trajectory and its flow time, conditioned on visual features, proprioception, and language. During training, action tokens stochastically attend to either the full conditioning video or only its current frame. This conditioning stream is separate from the noisy video prediction branch and may receive noise augmentation while preserving the current frame. Visual conditioning is disabled for samples without video. At deployment, WB-WAM uses the current-frame pathway to denoise the action trajectory directly, without synthesizing future video. The joint objective combines video and action flow matching:

\begin{equation}
    \label{eq:joint_objective}
    \mathcal{L}(\boldsymbol{\theta};\mathcal{D})
    = \lambda_{\mathrm{vid}}\mathcal{L}_{\mathrm{vid}}
    + \lambda_{\mathrm{act}}\mathcal{L}_{\mathrm{act}}.
\end{equation}

Here, $\mathcal{L}_{\mathrm{vid}}$ and $\mathcal{L}_{\mathrm{act}}$ denote the expected flow-time-weighted mean squared errors between predicted and target flow velocities for video and action, respectively. The coefficients $\lambda_{\mathrm{vid}}$ and $\lambda_{\mathrm{act}}$ balance two objectives. The video loss is evaluated only on future latent frames of samples with video. The action loss is averaged over the full trajectory tensor, including unavailable and padded entries assigned zero target velocity by \autoref{eq:action_targets}. Optimization on $\mathcal{D}_{\mathrm I}$ yields $\boldsymbol{\theta}_{\mathrm I}$ which initializes task adaptation for both simulation and real-world comparisons.

\subsection{Stage II: Mid-training with Retargeted PICO Motion}

Mid-training adapts the pre-trained model to egocentric demonstrations represented in the shared physical action space. This stage uses PICO demonstrations exclusively, pairing egocentric observations with retargeted body, root, and hand references. For the tasks evaluated in the mid-training experiment, the PICO dataset includes human demonstrations of the same tasks subsequently learned from robot demonstrations in Stage III. Starting from $\boldsymbol{\theta}_{\mathrm I}$, joint video and action learning continues under \autoref{eq:joint_objective} to obtain $\boldsymbol{\theta}_{\mathrm{II}}$. The reference construction pipeline is described in \autoref{sec:pico_data}. These references provide kinematic supervision derived from human demonstrations; dynamic stability and contact feasibility on the physical robot are not established by retargeting alone.

\subsection{Stage III: Task-Specific Real-Robot Post-training}

For each robot task $k$, an independent model $\boldsymbol{\theta}_{\mathrm{III}}^k$ is initialized from $\boldsymbol{\theta}_{\mathrm{II}}$ and adapted using the teleoperation dataset $\mathcal{D}_R^k$. Inspired by the body tracking objectives in BeyondMimic~\cite{liao2026beyondmimic}, we augment the joint video and action objective with a differentiable forward kinematics (FK) loss. Body joint configurations are reconstructed from the predicted action flow and denormalized before FK evaluation. The loss penalizes position and orientation errors of selected G1 links in the pelvis frame:

\begin{equation}
    \label{eq:stage3_objective}
    \mathcal{L}_{\mathrm{III}}
    = \mathcal{L}(\boldsymbol{\theta};\mathcal{D}_R^k)
    + \lambda_{\mathrm{FK}}
    \left(
    \mathcal{L}_{\mathrm{pos}}
    + \beta_{\mathrm{rot}}\mathcal{L}_{\mathrm{rot}}
    \right),
\end{equation}

where $\mathcal{L}_{\mathrm{pos}}$ and $\mathcal{L}_{\mathrm{rot}}$ measure position and orientation discrepancies over valid reference targets.

\section{WB-Datasets}
\label{sec:datasets}

WB-WAM combines heterogeneous external data with two complementary collections of household demonstrations. The external sources provide broad coverage for Stage I. Our WB-Datasets comprise PICO recordings reserved for motion transfer in Stage II and robot teleoperation demonstrations for task adaptation in Stage III. The PICO collection contains 22 hours of egocentric human demonstrations paired with body, root, and hand references for the G1 platform with Wuji dexterous hands. The robot collection contains 1,011 episodes spanning eight task groups, corresponding to 3.37 hours of recorded samples. This section describes the external data curation and the two acquisition pipelines.

\subsection{Heterogeneous Pre-training Data}
\label{sec:external_pre-training_data}

\begin{figure}[t]
    \centering
    \vspace{1mm}
    \includegraphics[width=0.96\linewidth]{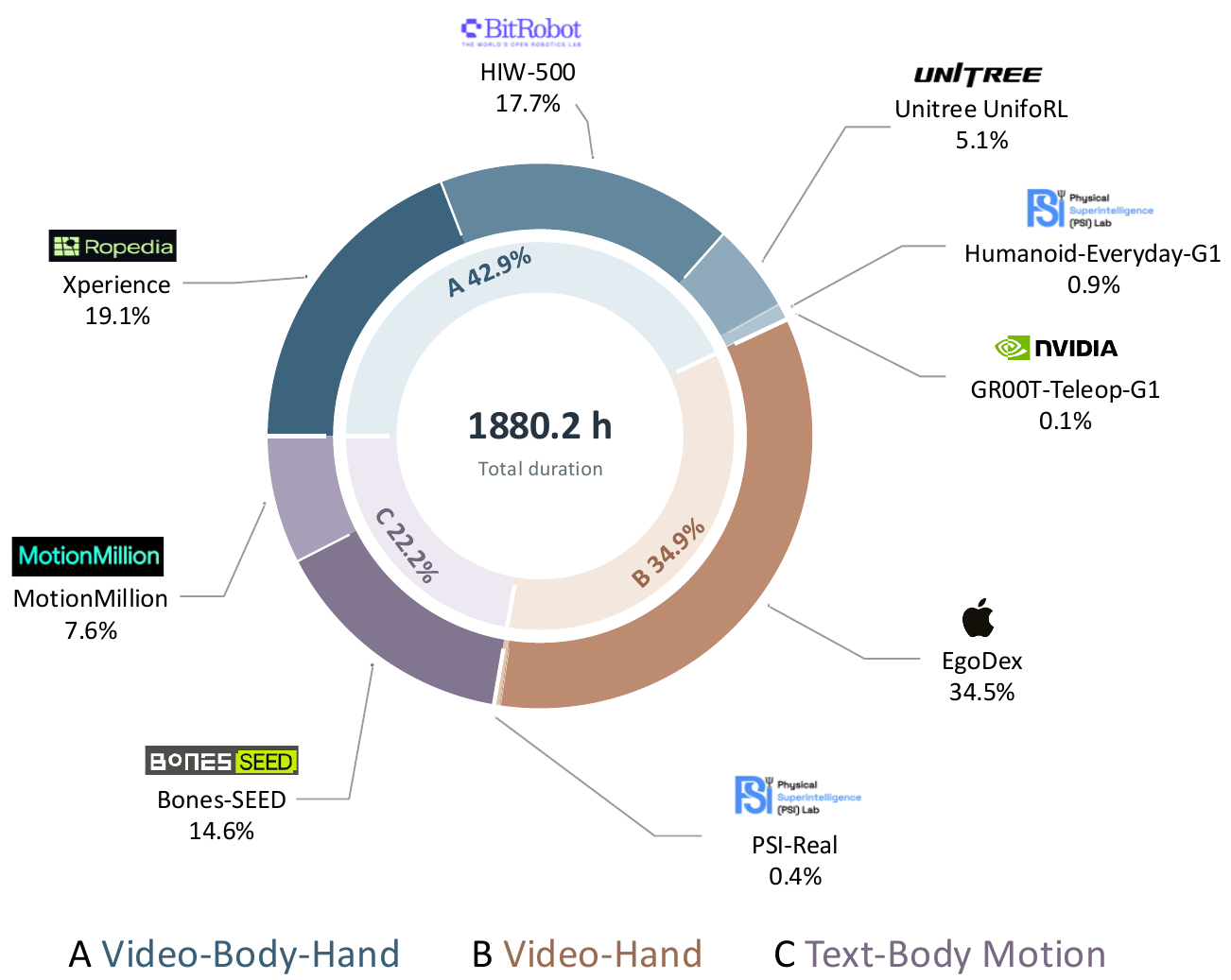}
    \caption{Composition of 9-source dataset used for pre-training.}
    \label{fig:datasets_overview}
    \vspace{-5mm}
\end{figure}

As shown in \autoref{fig:datasets_overview}, stage I uses 1880.2 hours of heterogeneous data from nine external sources. Video-based sources include Xperience-10M~\cite{ropedia2026xperience10m},
EgoDex~\cite{hoque2026egodex}, HIW-500~\cite{hiw500_2026}, Humanoid Everyday~\cite{zhao2025humanoideveryday}, GR00T-Teleop-G1~\cite{nvidia2025gr00tteleopg1}, and the Unitree and PSI-Real~\cite{wei2026psi_0} collections. Motion-only sources include MotionMillion~\cite{fan2025go} and BONES-SEED~\cite{studio142bones}.

Annotations are converted to the temporal and action interfaces used by WB-WAM. For Xperience-10M and MotionMillion, General Motion Retargeting (GMR)~\cite{araujo2025retargeting} retargets human motion to the G1 morphology. We then retarget Xperience-10M and EgoDex from human hands to wuji hands joint. Data already represented in robot coordinates undergo schema conversion and coordinate alignment without additional retargeting. Available body, root, and hand annotations populate their respective channels in \autoref{eq:physical_reference}, with unavailable modalities and action fields recorded in the metadata and validity masks.

\begin{figure}[t]
    \centering
    \vspace{1mm}
    \includegraphics[width=\columnwidth]{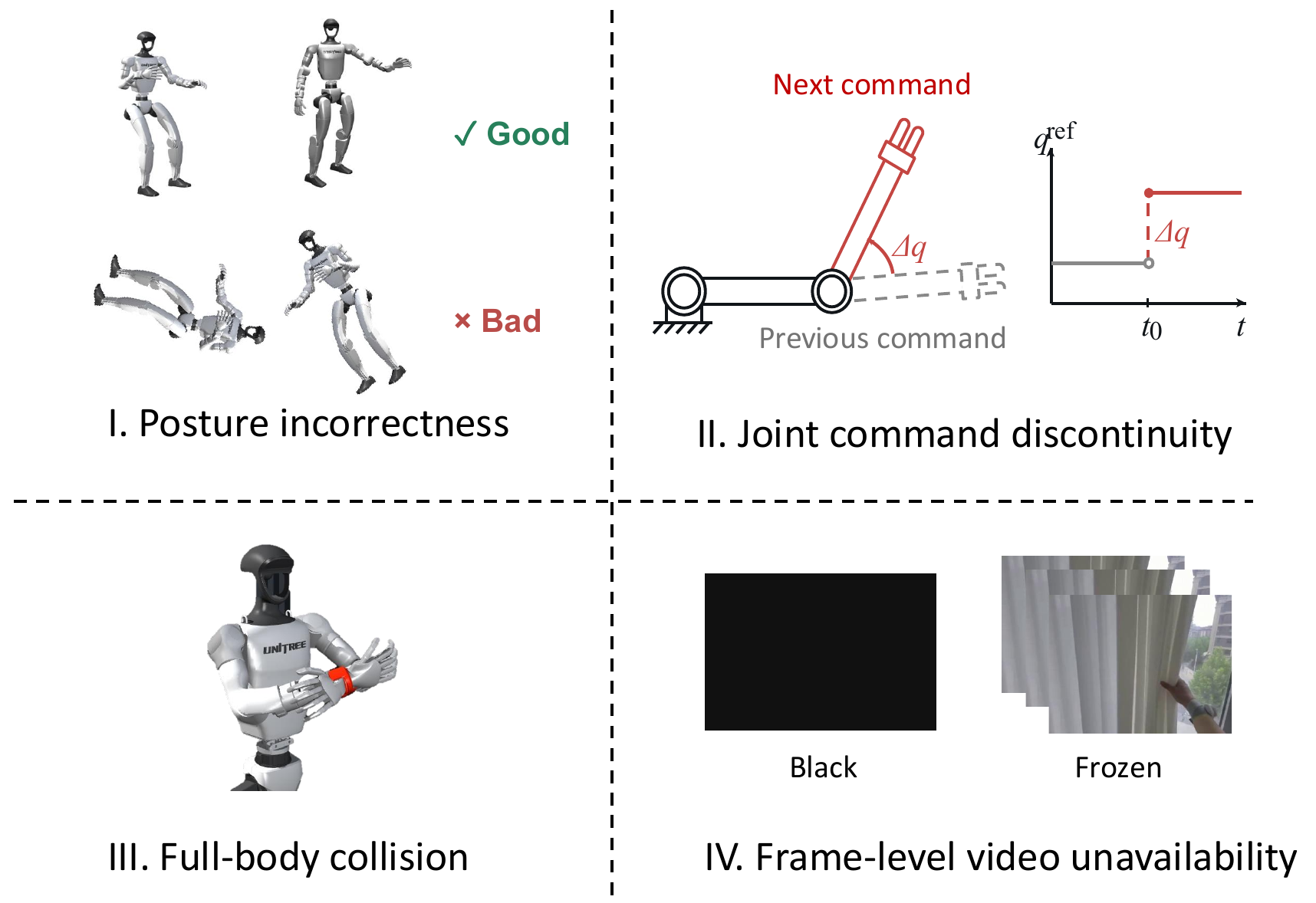}
    \caption{Pre-training data curation with pose validity,
    motion continuity, collision, and video quality checks.}
    \label{fig:data_process}
    \vspace{-5mm}
\end{figure}

The data processing pipeline checks for invalid poses, motion discontinuities, body or hand collisions, and missing or corrupted video frames (\autoref{fig:data_process}). Thresholded detector scores distinguish clear failures from ambiguous cases, which receive more detailed manual inspection. The processed pre-training data contains 1,880.2 hours of data in total.
% Figure 4

\subsection{PICO Egocentric Mid-training Data}
\label{sec:pico_data}

Each collector uses a PICO 4 Ultra and five trackers attached to the hands, feet, and waist. The setup records egocentric RGB video and synchronized SMPL body motion at 20~Hz. An initial G1 motion sequence is obtained using GMR and refined through constrained whole-body inverse kinematics. The refinement tracks calibrated human task-space position and palm-orientation targets while accounting for joint limits, velocity bounds, support constraints, and self-collision avoidance. The resulting body and root references are temporally aligned with the recorded video.

Dexterous hand references are obtained by reconstructing human hand keypoints from egocentric video with MINT~\cite{zhu2026mint} and retargeting to Wuji joint configurations.

The PICO recordings comprise 13,396 episodes across 73 tasks. For the mid-training study, PICO and robot demonstrations share the same task objectives, providing task-aligned human and robot data for evaluating transfer with limited robot supervision.

\subsection{SONIC Real-Robot Post-training Data}

Real-robot demonstrations are collected using the SONIC whole-body teleoperation system~\cite{luo2026sonic}. Body and root references are provided through a PICO 4 Ultra and a five-point tracking system, while MANUS gloves control the two Wuji hands. The collection covers eight tasks:
\textit{wipe the table}, \textit{close the curtain},
\textit{make the bed}, \textit{move the pillow},
\textit{push the cart}, \textit{checkout},
\textit{tidy the cloth}, and \textit{pick and place fruit}.
The fruit task includes three object categories: apple, lemon, and orange. The raw dataset contains 1,011 episodes and 242,668 frames at 20~Hz, corresponding to 3.37 hours of recorded sample coverage. 

\section{Experiments}

To investigate how humanoid robots learn loco-manipulation skills from whole-body heterogeneous data and adapt to real-world tasks with fewer robot demonstrations, we address three questions concerning WB-WAM's effectiveness, the contribution of whole-body pre-training, and the data efficiency of PICO mid-training:

\begin{itemize}
    \item \textbf{Q1: How does WB-WAM compare with existing
policies on humanoid loco-manipulation tasks?} 
    \item \textbf{Q2: Does whole-body humanoid pre-training improve downstream loco-manipulation performance?} 
    \item \textbf{Q3: Can task-aligned PICO mid-training reduce the robot demonstrations required for downstream adaptation?} 
\end{itemize}

\begin{figure}[!t]
    \centering
    \vspace{1mm}
    \includegraphics[width=0.98\linewidth]
    {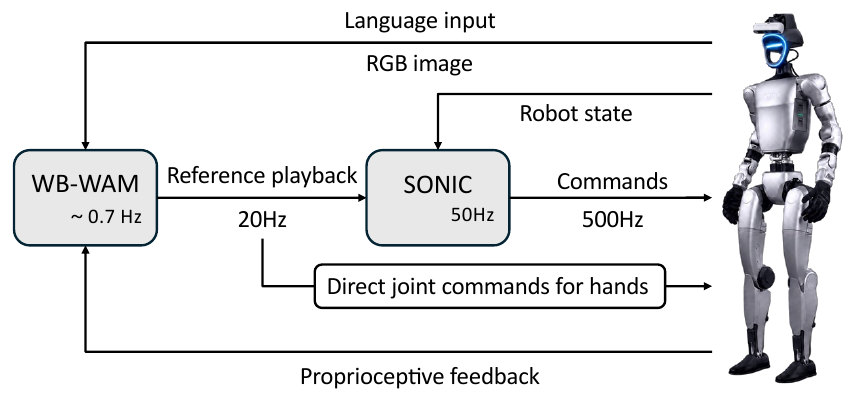}
    \caption{Robot hardware and system.}
    \label{fig:system}
    \vspace{-6mm}
\end{figure}

\subsection{Training Details}
The video expert is initialized from Wan2.2~\cite{wan2025wan}. The action expert uses a reduced-width DiT initialized through structural parameter transfer from the video backbone. The action input and output projections and the proprioceptive encoder are randomly initialized. The action output head produces a 96-D vector, with at most 72-D used and the remainder reserved for future use. Across all three stages, the video expert, action expert, and proprioceptive encoder are jointly optimized, while the video VAE and UMT5 text encoder remain frozen.

We pre-train the model for 1 epoch with a batch size of 16, using a sampling ratio of $3{:}1{:}1$ for video-body-hand, video-hand, and text-body-motion data. Following Fast-WAM~\cite{yuan2026fast}, we train the action branch to condition on either the full video or only its first frame with equal probability. This enables the same model to predict actions from generated future videos when IDM is enabled, or directly from the current observation when IDM is disabled, reducing inference cost.

For simulation experiments, WB-WAM is adapted to the observation and action interface used by the official HumanoidArena~\cite{wang2026humanoidarena} baselines. For the five-task real-world comparison, all six baseline policies are trained for 50 epochs on the same task-specific robot demonstrations and evaluated through a common execution interface. For the PICO transfer experiment, a shared $\boldsymbol{\theta}_{\mathrm I}$ is mid-trained for 10 epochs on the full PICO dataset, comprising 22 hours and 13,396 episodes across 73 tasks. The resulting checkpoint initializes task-specific policies, each post-trained for 50 epochs using 30 robot demonstrations.

\subsection{Deployment and System Details}
As illustrated in \autoref{fig:system}, WB-WAM is deployed on a Unitree G1 equipped with two Wuji dexterous hands and an Intel RealSense D455 camera. At each replanning step, the model receives the RGB observation, language instruction, proprioceptive state, and predicts a 32-timestep action chunk using 20 denoising steps. The first 20 action steps are executed at 20~Hz before replanning from updated observations.

The predicted action chunk is denormalized into physical references. At each playback step, body and root references are assembled into a ten-frame motion window sampled at 0.1~s intervals, starting from the current reference step. The frozen SONIC G1 encoder~\cite{luo2026sonic} maps this window into a 64-D controller latent, which conditions the tracking policy together with robot-state feedback. The tracking policy operates at 50~Hz, while a separate publishing loop sends low-level motor commands at 500~Hz. Hand references bypass the encoder and are issued directly as joint commands.

\subsection{Evaluation Metrics}
\label{sec:exp_metrics}

% Each quantitative model-task condition is evaluated in 20 real-robot trials. Final task success is the primary metric, while intermediate milestones are reported to distinguish failures in locomotion, object interaction, and task completion.

% \textit{Approach success (A)} indicates that the robot reaches the
% task-specific operating region from which interaction can begin. We
% report this metric for closing the curtain, wiping the table, and making the bed, where the robot must first walk to an appropriate
% interaction position.

% \textit{Contact success (C)} indicates that the robot establishes the
% intended functional contact with the task object. Depending on the task, this may require a valid grasp rather than incidental contact. For example, the robot must grasp the cart handle before pushing the cart.

% \textit{Locomotion success (L)} is additionally reported for cart
% pushing and measures successful execution of the locomotion stage while interacting with the cart.

% \textit{Task success (S)} indicates completion of the prescribed task
% goal. Intermediate metrics are evaluated independently over all trials rather than conditioned on success at an earlier milestone. When reporting the mean performance over multiple tasks, we average both the final task success rates (SR) and task progress (TP).

Each real-robot policy is evaluated in 20 trials per task. We report completion rates for approach (A), functional object contact (C), locomotion during cart pushing (L), and final task success (S), as applicable. Success rate (SR) measures final task completion, while task progress (TP) averages the milestone completion rates specified for each task in the tables, including S.
All milestone rates are computed over all trials. Mean SR and TP are obtained by averaging equally across tasks.

\begin{table*}[!t]
    \centering
    \vspace{2mm}
    \caption{
        Comparison on HumanoidArena~\cite{wang2026humanoidarena}.
        For each task, we report the strongest baseline from the original benchmark.
    }
    \label{tab:humanoidarena}

    \small
    \setlength{\tabcolsep}{4pt}
    \renewcommand{\arraystretch}{1.12}

    \resizebox{\textwidth}{!}{%
    \begin{tabular}{l*{7}{c}cc}
        \toprule
        Method
        & Football
        & DoubleDesk
        & P\&PBox
        & OpenDoor
        & SitSofa
        & Boxing
        & VisNavi
        & \multicolumn{2}{c}{Mean SR} \\
        \midrule

        \multirow{2}{*}{\shortstack[l]{Best baseline\\(per task)}}
        & DP
        & $\pi_{0.5}$
        & DP
        & DP
        & DP
        & DP
        & FM
        & FM 45.5$\pm$25.2\%
        & ACT 47.6$\pm$24.2\%
        \\

        & 45.0$\pm$10.8\%
        & 43.3$\pm$6.2\%
        & 75.0$\pm$4.1\%
        & 85.0$\pm$10.8\%
        & 78.3$\pm$14.3\%
        & 76.7$\pm$2.4\%
        & 38.3$\pm$14.3\%
        & $\pi_{0.5}$ 51.2$\pm$24.8\%
        & DP 60.0$\pm$24.2\%
        \\
        \midrule

        \rowcolor[gray]{0.90}
        \textbf{WB-WAM}
        & \textbf{70.0$\pm$8.2\%}
        & \textbf{65.0$\pm$4.1\%}
        & \textbf{86.7$\pm$2.4\%}
        & \textbf{98.3$\pm$2.4\%}
        & \textbf{95.0$\pm$4.1\%}
        & \textbf{81.7$\pm$2.4\%}
        & \textbf{76.7$\pm$4.7\%}
        & \multicolumn{2}{c}{\textbf{81.9$\pm$12.3\%}}
        \\
        \bottomrule
    \end{tabular}%
    }

    \vspace{-3mm}
\end{table*}

\begin{table*}[!t]
\vspace{2mm}
\centering
\caption{
Main comparison with 20 real-robot trials per task.
WB-WAM achieves the highest mean SR and mean TP.
}
\label{tab:exp_main}
\small
\setlength{\tabcolsep}{5.0pt}
\renewcommand{\arraystretch}{1.12}
\resizebox{\textwidth}{!}{%
\begin{tabular}{l*{15}{c}}
\toprule
& \multicolumn{9}{c}{w/ locomotion}
& \multicolumn{4}{c}{w/o locomotion}
& & \tabularnewline
\cmidrule(lr){2-10}\cmidrule(lr){11-14}
Method
& \multicolumn{3}{c}{Wipe the table}
& \multicolumn{3}{c}{Close the curtain}
& \multicolumn{3}{c}{Make the bed}
& \multicolumn{2}{c}{Move the pillow}
& \multicolumn{2}{c}{Tidy the cloth}
& Mean SR
& Mean TP \tabularnewline
\cmidrule(lr){2-4}\cmidrule(lr){5-7}\cmidrule(lr){8-10}
\cmidrule(lr){11-12}\cmidrule(lr){13-14}
& A & C & S
& A & C & S
& A & C & S
& C & S
& C & S
& & \tabularnewline
\midrule
ACT~\cite{zhao2023learning}
& 0/20 & 0/20 & 0/20
& 0/20 & 0/20 & 0/20
& 0/20 & 0/20 & 0/20
& 16/20 & \underline{8/20}
& 19/20 & 12/20
& 20\%
& 27.5\% \tabularnewline
% Diffusion Policy~\cite{chi2025diffusion}
% & 0/20 & 0/20 & 0/20
% & 0/20 & 0/20 & 0/20
% & 0/20 & 0/20 & 0/20
% & 0/20 & 0/20
% & 0/20 & 0/20
% & 0\%
% & 0\% \tabularnewline
$\pi_{0.5}$~\cite{intelligence2025pi_}
& 13/20 & 12/20 & 12/20
& 11/20 & 10/20 & 6/20
& 11/20 & 10/20 & 10/20
& 19/20 & \underline{8/20}
& 16/20 & 16/20
& 52\%
& 61.17\% \tabularnewline
GR00T N1.6~\cite{nvidia2025gr00t}
& 16/20 & 15/20 & \underline{15/20}
& 18/20 & 13/20 & \underline{12/20}
& 0/20 & 0/20 & 0/20
& 6/20 & 0/20
& 17/20 & 15/20
& 42\%
& 48.67\% \tabularnewline
Fast-WAM~\cite{yuan2026fast}
& 0/20 & 0/20 & 0/20
& 8/20 & 3/20 & 2/20
& 0/20 & 0/20 & 0/20
& 4/20 & 0/20
& 9/20 & 4/20
& 6\%
& 12.83\% \tabularnewline
DiT4DiT~\cite{ma2026dit4dit}
& 12/20 & 6/20 & 3/20
& 16/20 & 15/20 & 7/20
& 16/20 & 9/20 & 9/20
& 5/20 & 0/20
& 16/20 & 12/20
& 31\%
& 47.5\% \tabularnewline
OpenWAM~\cite{wang2026openwam}
& 18/20 & 18/20 & \textbf{18/20}
& 11/20 & 10/20 & 8/20
& 18/20 & 17/20 & \underline{17/20}
& 20/20 & \textbf{20/20}
& 17/20 & \underline{17/20}
& \underline{80\%}
& \underline{82\%} \tabularnewline
\midrule
\rowcolor[gray]{0.90}
\textbf{WB-WAM}
& 16/20 & 16/20 & \underline{15/20}
& 16/20 & 15/20 & \textbf{13/20}
& 19/20 & 18/20 & \textbf{18/20}
& 20/20 & \textbf{20/20}
& 18/20 & \textbf{18/20}
& \textbf{84\%}
& \textbf{86.67\%} \tabularnewline
% \rowcolor[gray]{0.90}
% \textbf{WB-WAM (w/ PICO)}
% & 19/20 & 18/20 & \textbf{18/20}
% & 19/20 & 17/20 & \textbf{16/20}
% & -- & -- & --
% & -- & --
% & -- & --
% & --
% & -- \tabularnewline
\bottomrule
\end{tabular}%
}
\vspace{-3mm}
\end{table*}

\subsection{Main Comparison}
We evaluate WB-WAM against representative imitation-learning, vision-language-action, and WAM baselines to examine its benefits for humanoid loco-manipulation. We first evaluate coordinated whole-body behavior in simulation, then investigate whether the observed advantages persist on a physical humanoid.

\subsubsection{Simulation Benchmark Evaluation}
\label{sec:exp_simulation}
We first evaluate WB-WAM on HumanoidArena~\cite{wang2026humanoidarena},
which comprises three human-object interaction tasks and four
human-scene interaction tasks requiring coordinated whole-body
behavior. We follow the in-GMT protocol, using SONIC for both
demonstration collection and policy execution. Success is assessed
using the benchmark-defined task criteria, and comparisons are
made against the highest reported baseline success rate for each
task under the same SONIC setting.

Across the seven HumanoidArena tasks, WB-WAM achieves
81.9\% mean success and significantly outperforms the strongest
reported baseline on all seven tasks
(\autoref{tab:humanoidarena}). These gains span locomotion,
posture adjustment, and object interaction, demonstrating the
effectiveness of whole-body-pre--trained WB-WAM across diverse
humanoid behaviors. We next examine whether these advantages
persist during real-world task execution.

\subsubsection{Real-World Comparison}
\label{sec:exp_comparison}
We next examine whether the advantages observed in simulation
persist during physical task execution, where successful
manipulation requires accurate approach, contact establishment,
and coordinated body motion.

We compare WB-WAM with six baselines:
ACT~\cite{zhao2023learning}, $\pi_{0.5}$~\cite{intelligence2025pi_}, GR00T
N1.6~\cite{nvidia2025gr00t}, Fast-WAM~\cite{yuan2026fast} (from Wan2.2), DiT4DiT~\cite{ma2026dit4dit}, and
OpenWAM~\cite{wang2026openwam}.

All methods are evaluated on five real-robot tasks: \textit{close the curtain, wipe the table, make the bed, move the pillow, and tidy the cloth}. The first three tasks require locomotion before or during manipulation, whereas the latter two focus on stationary whole-body manipulation. Every method uses the same set of approximately 100 robot demonstrations per task. WB-WAM is initialized from $\boldsymbol{\theta}_{\mathrm I}$.

\begin{figure}[!t]
    \centering
    \vspace{1mm}
    \includegraphics[width=0.9\linewidth]
    {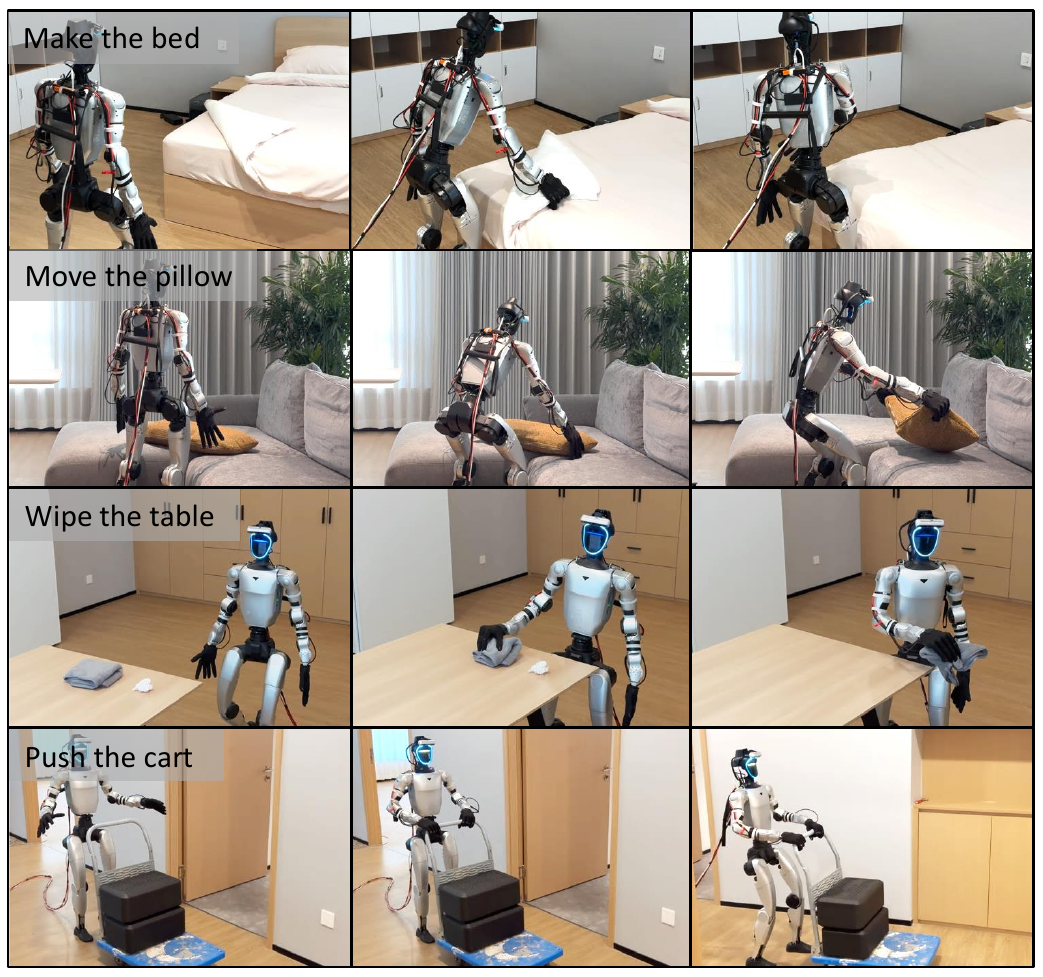}
    \caption{Real-world experiments. WB-WAM enables the robot to perform a variety of whole-body manipulation tasks.}
    \label{fig:Exp}
    \vspace{-6mm}
\end{figure}

The real-world results show a consistent advantage
(\mbox{\autoref{tab:exp_main}}). Across the three tasks requiring locomotion, WB-WAM achieves 76.7\% mean success, compared with 71.7\% for OpenWAM, the strongest evaluated baseline. Across all five tasks, mean success reaches 84.0\%, compared with 80.0\% for OpenWAM. Baseline failures frequently involve inaccurate approach, failed grasping, or poor coordination between manipulation and body motion. Together, the simulation and real-world evaluations support whole-body action pre-training as an effective foundation for humanoid loco-manipulation.

\subsection{PICO Mid-training}
\label{sec:exp_pico}

\begin{table*}[t]
\centering
\vspace{2mm}
\caption{
Effect of PICO mid-training with 20 real-robot trials per condition.
}
\label{tab:exp_pico}
\small
\setlength{\tabcolsep}{5.0pt}
\renewcommand{\arraystretch}{1.12}
\resizebox{\textwidth}{!}{%
\begin{tabular}{l*{14}{c}}
\toprule
& & \multicolumn{9}{c}{w/ locomotion}
& \multicolumn{2}{c}{\makebox[0pt][c]{w/o locomotion}}
& & \tabularnewline
\cmidrule(lr){3-11}\cmidrule(lr){12-13}
\multirow{2}{*}[-0.6ex]{Training route}
& \multirow{2}{*}[-0.6ex]{\shortstack{Robot\\[-1pt]Demos}}
& \multicolumn{3}{c}{Close the curtain}
& \multicolumn{3}{c}{Push the cart}
& \multicolumn{3}{c}{Wipe the table}
& \multicolumn{2}{c}{Checkout}
& \multirow{2}{*}[-0.6ex]{Mean SR}
& \multirow{2}{*}[-0.6ex]{Mean TP} \tabularnewline
\cmidrule(lr){3-5}\cmidrule(lr){6-8}
\cmidrule(lr){9-11}\cmidrule(lr){12-13}
&
& A & C & S
& C & L & S
& A & C & S
& C & S
& & \tabularnewline
\midrule
Direct post-training
& 30
& 13/20 & 12/20 & 12/20
& 13/20 & 11/20 & \underline{11/20}
& 12/20 & 11/20 & 11/20
& 6/20 & 5/20
& 48.75\%
& 51.04\% \tabularnewline
Direct post-training
& 100
& 16/20 & 15/20 & \underline{13/20}
& 19/20 & 16/20 & \textbf{16/20}
& 16/20 & 16/20 & \underline{15/20}
& 10/20 & \underline{8/20}
& \underline{65\%}
& \underline{70.42\%} \tabularnewline
\midrule
\rowcolor[gray]{0.90}
\textbf{w/ PICO mid-training}
& 30
& 19/20 & 17/20 & \textbf{16/20}
& 16/20 & 17/20 & \textbf{16/20}
& 19/20 & 18/20 & \textbf{18/20}
& 12/20 & \textbf{9/20}
& \textbf{73.75\%}
& \textbf{78.13\%} \tabularnewline
\bottomrule
\end{tabular}%
}
\vspace{-3mm}
\end{table*}

% We next evaluate whether PICO mid-training improves adaptation efficiency on downstream robot tasks. The evaluation includes four tasks covered by the PICO data: \textit{closing the curtain, pushing the cart, checkout, and wiping the table.} All configurations originate from $\boldsymbol{\theta}_{\mathrm I}$.

% PICO mid-training offers a practical way to reduce data collection costs by leveraging PICO demonstrations that are easier to collect than real-robot demonstrations. We use approximately 150 PICO demonstrations per task for joint mid-training, followed by post-training with only 30 real-robot demonstrations per task. The resulting policy achieves \textbf{73.8\%} mean success, exceeding direct post-training with 100 robot demonstrations (65.0\%).

We next evaluate whether PICO mid-training improves adaptation
efficiency on four downstream robot tasks. All
configurations originate from $\boldsymbol{\theta}_{\mathrm I}$. Mid-training uses the full 73-task PICO dataset, which
includes approximately 150 PICO demonstrations for each of
the four evaluated tasks. A separate policy for each task
is then initialized from the same shared $\boldsymbol{\theta}_{\mathrm {II}}$
and post-trained using 30 real-robot demonstrations. 

As shown in \autoref{tab:exp_pico}, the resulting task-specific
policies achieve 73.8\% mean success, exceeding direct
post-training with 100 robot demonstrations per task (65.0\%). This reduces the number of robot demonstrations by 70\% while improving mean success. At the same 30-demonstration robot-data budget, direct post-training achieves only 48.8\% mean success, further demonstrating the benefit of PICO mid-training. On the two tasks shared with \autoref{tab:exp_main}, PICO mid-training matches the best baseline success rate on table wiping (90\%) and exceeds it on curtain closing (80\% vs.\ 60\%), using only 30 robot demonstrations per task compared with approximately 100 for the baselines.

\subsection{Language-Conditioned Manipulation}
\label{sec:exp_language}

Finally, we evaluate language-conditioned fruit selection using a single WB-WAM policy trained on approximately 300 demonstrations spanning orange, lemon, and apple. The results in \autoref{tab:exp_language} show that the shared policy can select the instructed fruit and complete pick-and-place without training separate policies for different targets.

\begin{table}[t]
\centering
\caption{
Language-conditioned fruit manipulation.
}
\label{tab:exp_language}
\small
\setlength{\tabcolsep}{3.5pt}
\renewcommand{\arraystretch}{1.12}
\begin{tabular}{l*{6}{c}}
\toprule
& \multicolumn{2}{c}{Orange}
& \multicolumn{2}{c}{Lemon}
& \multicolumn{2}{c}{Apple} \tabularnewline
\cmidrule(lr){2-3}\cmidrule(lr){4-5}\cmidrule(lr){6-7}
Method
& C & S
& C & S
& C & S \tabularnewline
\midrule
\rowcolor[gray]{0.90}
\textbf{WB-WAM}
& 17/20 & 15/20
& 18/20 & 17/20
& 19/20 & 16/20 \tabularnewline
\bottomrule
\end{tabular}
\vspace{-5mm}
\end{table}

\subsection{Visuomotor Generalization}
\label{sec:visual_generalization}

\begin{figure}[!htbp]
    \centering
    % \vspace{-2mm}
    \includegraphics[width=\linewidth]
    {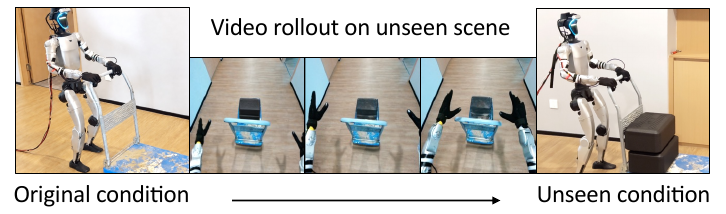}
    \caption{\textbf{Video prediction and real-world execution
    under visual variation.}
    The three central frames show a video rollout generated by the video expert under a modified visual condition. Real-robot cart pushing is shown in a familiar scene (left) and a novel scene absent from the dataset (right).}
    \label{fig:visual_generalization}
    \vspace{-4mm}
\end{figure}

We qualitatively evaluate WB-WAM on cart pushing under an unseen visual condition (\autoref{fig:visual_generalization}). Additional objects alter the appearance of the cart while the task objective remains unchanged. The video expert can optionally roll out future visual observations under the modified condition, while WB-WAM executes the task on the real robot without further adaptation, illustrating visuomotor generalization.

\section{Conclusion}

This work presents WB-WAM, a World Action Model that integrates heterogeneous body and hand supervision into generative video pre-training for humanoid loco-manipulation. A shared physical action space connects broad pre-training, retargeted PICO motion, and task-specific robot adaptation. Simulation benchmarks and real-world experiments jointly
demonstrate the effectiveness of whole-body-pre-trained
WB-WAM for humanoid loco-manipulation. Task-aligned PICO mid-training further improves mean success with few robot demonstrations per task, highlighting the value of human motion for reducing robot data requirements. The current evaluation is limited to one robot platform and task-specific policies. Extending transfer to unseen tasks and improving execution accuracy during object interaction are important directions for future work.


%%%%%%%%%%%%%%%%%%%%%%%%%%%%%%%%%%%%%%%%%%%%%%%%%%%%%%%%%%%%%%%%%%%%%%%%%%%%%%%%

%%%%%%%%%%%%%%%%%%%%%%%%%%%%%%%%%%%%%%%%%%%%%%%%%%%%%%%%%%%%%%%%%%%%%%%%%%%%%%%%



\section*{ACKNOWLEDGMENT}
This work uses BONES-SEED motion data for heterogeneous
pre-training. Motion Data by Bones Studio -
\url{https://bones.studio/}. OpenAI ChatGPT was used to generate selected illustrative elements: the circular graphic and purple human figure in Fig.~1; the purple human figure and proprioceptive-state illustration in Fig.~2; the circular graphic in Fig.~3; and the joint-command-discontinuity illustration in Fig.~4 (II).

%%%%%%%%%%%%%%%%%%%%%%%%%%%%%%%%%%%%%%%%%%%%%%%%%%%%%%%%%%%%%%%%%%%%%%%%%%%%%%%%

% References are important to the reader; therefore, each citation must be complete and correct. If at all possible, references should be commonly available publications.

% \begin{thebibliography}{99}

% \bibitem{c1} G. O. Young, ÒSynthetic structure of industrial plastics (Book style with paper title and editor),Ó 	in Plastics, 2nd ed. vol. 3, J. Peters, Ed.  New York: McGraw-Hill, 1964, pp. 15-64.
% \bibitem{c2} W.-K. Chen, Linear Networks and Systems (Book style).	Belmont, CA: Wadsworth, 1993, pp. 123-135.

\bibliographystyle{IEEEtran}
\bibliography{bibtex/main}

@article{zhao2023learning,
  title={Learning fine-grained bimanual manipulation with low-cost hardware},
  author={Zhao, Tony Z and Kumar, Vikash and Levine, Sergey and Finn, Chelsea},
  journal={arXiv preprint arXiv:2304.13705},
  year={2023}
}

@article{intelligence2025pi_,
  title   = {{$\pi_{0.5}$: a Vision-Language-Action Model with Open-World Generalization}},
  author  = {{Physical Intelligence} and Black, Kevin and Brown, Noah
             and Darpinian, James and Dhabalia, Karan and Driess, Danny
             and Esmail, Adnan and Equi, Michael and Finn, Chelsea
             and Fusai, Niccolo and others},
  journal = {arXiv preprint arXiv:2504.16054},
  year    = {2025}
}

@misc{nvidia2025gr00t,
  title={Gr00t n1.6: An improved open foundation model for generalist humanoid robots},
  author={NVIDIA GEAR Team and Azzolini, Allison and Bjorck, Johan and Blukis, Valts and others},
  year={2025}
}

@article{bjorck2025gr00t,
  title={Gr00t n1: An open foundation model for generalist humanoid robots},
  author={Bjorck, Johan and Casta{\~n}eda, Fernando and Cherniadev, Nikita and Da, Xingye and Ding, Runyu and Fan, Linxi and Fang, Yu and Fox, Dieter and Hu, Fengyuan and Huang, Spencer and others},
  journal={arXiv preprint arXiv:2503.14734},
  year={2025}
}

@article{wei2026psi_0,
  title   = {{$\Psi_0$: An Open Foundation Model Towards Universal Humanoid Loco-Manipulation}},
  author  = {Wei, Songlin and Jing, Hongyi and Li, Boqian and Zhao, Zhenyu
             and Mao, Jiageng and Ni, Zhenhao and He, Sicheng and Liu, Jie
             and Liu, Xiawei and Kang, Kaidi and others},
  journal = {arXiv preprint arXiv:2603.12263},
  year    = {2026}
}

@article{hu2026openhlm,
  title={OpenHLM: An Empirical Recipe for Whole-Body Humanoid Loco-Manipulation},
  author={Hu, Yingdong and Zhu, Haodong and Zheng, Boyuan and Hu, Yihang and Zhang, Tong and Chen, Zunhao and Zhao, Junming and Nai, Ruiqian and Gao, Yang},
  journal={arXiv preprint arXiv:2606.22174},
  year={2026}
}

@inproceedings{jiang2026wholebodyvla,
  title={Wholebodyvla: Towards unified latent vla for whole-body loco-manipulation control},
  author={Jiang, Haoran and Chen, Jin and Bu, Qingwen and Chen, Li and Shi, Modi and Zhang, Yanjie and Li, Delong and Suo, Chuanzhe and Li, Hongyang and others},
  booktitle={International Conference on Learning Representations},
  volume={2026},
  pages={157438--157461},
  year={2026}
}

@article{shi2026egohumanoid,
  title={Egohumanoid: Unlocking in-the-wild loco-manipulation with robot-free egocentric demonstration},
  author={Shi, Modi and Peng, Shijia and Chen, Jin and Jiang, Haoran and Li, Tianyu and Huang, Di and Luo, Ping and Li, Hongyang and Chen, Li},
  journal={arXiv preprint arXiv:2602.10106},
  year={2026}
}

@article{luo2026sonic,
  title={Sonic: Supersizing motion tracking for natural humanoid whole-body control},
  author={Luo, Zhengyi and Yuan, Ye and Wang, Tingwu and Li, Chenran and Casta{\~n}eda, Fernando and Chen, Sirui and Cao, Zi-Ang and Li, Jiefeng and Minor, David and Ben, Qingwei and others},
  journal={Science Robotics},
  volume={11},
  number={117},
  pages={eaed4592},
  year={2026},
  publisher={American Association for the Advancement of Science}
}

@article{liao2026beyondmimic,
  title={Beyondmimic: From motion tracking to versatile humanoid control via guided diffusion},
  author={Liao, Qiayuan and Truong, Takara E and Huang, Xiaoyu and Gao, Yuman and Tevet, Guy and Sreenath, Koushil and Liu, C Karen},
  journal={Science Robotics},
  volume={11},
  number={117},
  pages={eadx8924},
  year={2026},
  publisher={American Association for the Advancement of Science}
}

@article{araujo2025retargeting,
  title={Retargeting matters: General motion retargeting for humanoid motion tracking},
  author={Araujo, Joao Pedro and Ze, Yanjie and Xu, Pei and Wu, Jiajun and Liu, C Karen},
  journal={arXiv preprint arXiv:2510.02252},
  year={2025}
}

@article{hu2024video,
  title={Video prediction policy: A generalist robot policy with predictive visual representations},
  author={Hu, Yucheng and Guo, Yanjiang and Wang, Pengchao and Chen, Xiaoyu and Wang, Yen-Jen and Zhang, Jianke and Sreenath, Koushil and Lu, Chaochao and Chen, Jianyu},
  journal={arXiv preprint arXiv:2412.14803},
  year={2024}
}

@article{pai2025mimicvideo,
  title={mimic-video: Video-action models for generalizable robot control beyond vlas},
  author={Pai, Jonas and Achenbach, Liam and Montesinos, Victoriano and Forrai, Benedek and Mees, Oier and Nava, Elvis},
  journal={arXiv preprint arXiv:2512.15692},
  year={2025}
}

@article{li2026causal,
  title={Causal world modeling for robot control},
  author={Li, Lin and Zhang, Qihang and Luo, Yiming and Yang, Shuai and Wang, Ruilin and Han, Fei and Yu, Mingrui and Gao, Zelin and Xue, Nan and Zhu, Xing and others},
  journal={arXiv preprint arXiv:2601.21998},
  year={2026}
}

@article{cen2025worldvla,
  title={Worldvla: Towards autoregressive action world model},
  author={Cen, Jun and Yu, Chaohui and Yuan, Hangjie and Jiang, Yuming and Huang, Siteng and Guo, Jiayan and Li, Xin and Song, Yibing and Luo, Hao and Wang, Fan and others},
  journal={arXiv preprint arXiv:2506.21539},
  year={2025}
}

@inproceedings{bi2026motus,
  title={Motus: A unified latent action world model},
  author={Bi, Hongzhe and Tan, Hengkai and Xie, Shenghao and Wang, Zeyuan and Huang, Shuhe and Liu, Haitian and Zhao, Ruowen and Feng, Yao and Xiang, Chendong and Rong, Yinze and others},
  booktitle={Proceedings of the IEEE/CVF Conference on Computer Vision and Pattern Recognition},
  pages={35101--35113},
  year={2026}
}

@article{kim2026cosmospolicy,
  title={Cosmos policy: Fine-tuning video models for visuomotor control and planning},
  author={Kim, Moo Jin and Gao, Yihuai and Lin, Tsung-Yi and Lin, Yen-Chen and Ge, Yunhao and Lam, Grace and Liang, Percy and Song, Shuran and Liu, Ming-Yu and Finn, Chelsea and others},
  journal={arXiv preprint arXiv:2601.16163},
  year={2026}
}

@inproceedings{li2026egowam,
  title={EgoWAM: World Action Models Beyond Pixels with In-the-Wild Egocentric Human Data},
  author={Li, Baoyu and Yin, Xinchen and Lin, Mengying and Zhang, Yixin and Xu, Danfei},
  booktitle={Robot World Models},
  year={2026}
}

@article{zhu2025unified,
  title={Unified world models: Coupling video and action diffusion for pretraining on large robotic datasets},
  author={Zhu, Chuning and Yu, Raymond and Feng, Siyuan and Burchfiel, Benjamin and Shah, Paarth and Gupta, Abhishek},
  journal={arXiv preprint arXiv:2504.02792},
  year={2025}
}

@article{ye2026world,
  title={World action models are zero-shot policies},
  author={Ye, Seonghyeon and Ge, Yunhao and Zheng, Kaiyuan and Gao, Shenyuan and Yu, Sihyun and Kurian, George and Indupuru, Suneel and Tan, You Liang and Zhu, Chuning and Xiang, Jiannan and others},
  journal={arXiv preprint arXiv:2602.15922},
  year={2026}
}

@article{ma2026dit4dit,
  title={Dit4dit: Jointly modeling video dynamics and actions for generalizable robot control},
  author={Ma, Teli and Zheng, Jia and Wang, Zifan and Jiang, Chunli and Cui, Andy and Liang, Junwei and Yang, Shuo},
  journal={arXiv preprint arXiv:2603.10448},
  year={2026}
}

@article{yuan2026fast,
  title={Fast-wam: Do world action models need test-time future imagination?},
  author={Yuan, Tianyuan and Dong, Zibin and Liu, Yicheng and Zhao, Hang},
  journal={arXiv preprint arXiv:2603.16666},
  year={2026}
}

@article{zhou2026tau_0,
  title   = {{$\tau_0$-WM: A Unified Video-Action World Model for Robotic Manipulation}},
  author  = {Zhou, Pengfei and Chen, Shengcong and Chen, Di and Wang, Jiaxu
             and Jin, Rongjun and Zhu, Bingwen and Pan, Yike and Gu, Songen
             and Wang, Kuanning and Nan, Shufeng and others},
  journal = {arXiv preprint arXiv:2606.01027},
  year    = {2026}
}

@article{wang2026openwam,
  title={OpenWAM: An Open, Modular Exploration Towards Systematic World-Action Model Pretraining},
  author={Wang, Yuran and Huang, Siqiao and Li, Mingleyang and Zhang, Chenhao and Liang, Jiaqi and Jin, Weiyang and Chen, Yue and Chi, Xuemin and Zhou, Donghao and Yu, Qize and others},
  journal={arXiv preprint arXiv:2609.07398},
  year={2026}
}

@article{zheng2026motionwam,
  title={MotionWAM: Towards Foundation World Action Models for Real-Time Humanoid Loco-Manipulation},
  author={Zheng, Jia and Ma, Teli and Fan, Yudong and Wang, Zifan and Yang, Shuo and Liang, Junwei},
  journal={arXiv preprint arXiv:2606.09215},
  year={2026}
}

@misc{beingbeyond2026beingm07,
  title={{Being-M0.7}: A Latent World-Action Model for Humanoid Robots},
  author={Yue, Junpeng and Li, Boyuan and Wang, Yuxuan and Wang, Zepeng and Fu, Yuhui and Xie, Feiyang and Zhang, Yu and Zhang, Jing and Wang, Jiangxing and Lu, Zongqing},
  year={2026},
  howpublished={BeingBeyond Technical Report},
  url={https://research.beingbeyond.com/being-m07/being-m07.pdf}
}

@article{li2026omega0,
  title   = {{$\omega$-0: A Latent Predictive World Action Model for Concurrent Humanoid Loco-Manipulation}},
  author  = {Li, Zhe and Zhang, Zhenzhe and Wei, Yangyang and Zhang, Wenjie
             and Yuan, Xichen and Zhi, Peiyuan and Li, Gen and Guo, Xinying
             and Gao, Fengjie and Yang, Jianfei and others},
  journal = {arXiv preprint arXiv:2608.06375},
  year    = {2026}
}

@article{wan2025wan,
  title={Wan: Open and advanced large-scale video generative models},
  author={Wan, Team and Wang, Ang and Ai, Baole and Wen, Bin and Mao, Chaojie and Xie, Chen-Wei and Chen, Di and Yu, Feiwu and Zhao, Haiming and Yang, Jianxiao and others},
  journal={arXiv preprint arXiv:2503.20314},
  year={2025}
}

@article{agarwal2025cosmos,
  title={Cosmos world foundation model platform for physical ai},
  author={Agarwal, Niket and Ali, Arslan and Bala, Maciej and Balaji, Yogesh and Barker, Erik and Cai, Tiffany and Chattopadhyay, Prithvijit and Chen, Yongxin and Cui, Yin and Ding, Yifan and others},
  journal={arXiv preprint arXiv:2501.03575},
  year={2025}
}

@article{zhu2026mint,
  title={MINT: A Unified Model for World-Space Camera and Hand Motion Estimation from Scalable Egocentric Pipeline Supervision},
  author={Zhu, Zijie and Cai, Weiren and Wang, Yizhou and Yang, Zhenjie and Liu, Yide and Chen, Jiahao and He, Guanqi},
  journal={arXiv preprint arXiv:2609.04958},
  year={2026}
}

@inproceedings{o2024open,
  title={Open x-embodiment: Robotic learning datasets and rt-x models: Open x-embodiment collaboration},
  author={O’Neill, Abby and Rehman, Abdul and Maddukuri, Abhiram and Gupta, Abhishek and Padalkar, Abhishek and Lee, Abraham and Pooley, Acorn and Gupta, Agrim and Mandlekar, Ajay and Jain, Ajinkya and others},
  booktitle={2024 IEEE International Conference on Robotics and Automation (ICRA)},
  pages={6892--6903},
  year={2024},
  organization={IEEE}
}

@article{black2024pi_0,
  title   = {{$\pi_0$: A Vision-Language-Action Flow Model for General Robot Control}},
  author  = {Black, Kevin and Brown, Noah and Driess, Danny and Esmail, Adnan
             and Equi, Michael and Finn, Chelsea and Fusai, Niccolo
             and Groom, Lachy and Hausman, Karol and Ichter, Brian and others},
  journal = {arXiv preprint arXiv:2410.24164},
  year    = {2024}
}

@inproceedings{hoque2026egodex,
  title={Egodex: Learning dexterous manipulation from large-scale egocentric video},
  author={Hoque, Ryan and Huang, Peide and Yoon, David and Zhang, Jian and others},
  booktitle={International Conference on Learning Representations},
  volume={2026},
  pages={4218--4237},
  year={2026}
}

@article{zhao2025humanoideveryday,
  title={Humanoid everyday: A comprehensive robotic dataset for open-world humanoid manipulation},
  author={Zhao, Zhenyu and Jing, Hongyi and Liu, Xiawei and Mao, Jiageng and Jha, Abha and Yang, Hanwen and Xue, Rong and Zakharov, Sergey and Guizilini, Vitor and Wang, Yue},
  journal={arXiv preprint arXiv:2510.08807},
  year={2025}
}

@misc{ropedia2026xperience10m,
  author       = {{Ropedia}},
  title        = {{Xperience-10M}: A Large-Scale Egocentric
                  Multimodal Dataset with Structured 3D/4D Annotations},
  year         = {2026},
  howpublished = {Hugging Face dataset},
  url          = {https://huggingface.co/datasets/ropedia-ai/xperience-10m}
}

@misc{hiw500_2026,
  title={HIW-500: Humanoids In-the-Wild Dataset for Robot Learning},
  author={BitRobot and Unitree and Hugging Face},
  year={2026},
  howpublished={\url{https://bitrobot-foundation.github.io/humanoids-in-the-wild-500-hours/}}
}

@inproceedings{fan2025go,
  title={Go to zero: Towards zero-shot motion generation with million-scale data},
  author={Fan, Ke and Lu, Shunlin and Dai, Minyue and Yu, Runyi and Xiao, Lixing and Dou, Zhiyang and Dong, Junting and Ma, Lizhuang and Wang, Jingbo},
  booktitle={2025 IEEE/CVF International Conference on Computer Vision (ICCV)},
  pages={13336--13348},
  year={2025},
  organization={IEEE}
}

@misc{studio142bones,
  title={BONES-SEED: Skeletal everyday embodiment dataset},
  author={Studio, Bones},
  year={2026},
  howpublished={\url{https://huggingface.co/datasets/bones-studio/seed/}}
}

@misc{nvidia2025gr00tteleopg1,
  title        = {Unitree G1 Fruits Pick and Place 1K Dataset},
  author       = {{NVIDIA GEAR}},
  year         = {2025},
  howpublished = {Hugging Face dataset},
  url          = {https://huggingface.co/datasets/nvidia/PhysicalAI-Robotics-GR00T-Teleop-G1}
}

@article{wang2026humanoidarena,
  title={HumanoidArena: Benchmarking Egocentric Hierarchical Whole-body Learning},
  author={Wang, Taowen and Xie, Zikang and Yang, Bin and Wang, Yunheng and Yuan, Zizhao and Fang, Yuetong and Feng, Yixiao and Wang, Yichi and Chen, Xingyu and Chen, Haodong and others},
  journal={arXiv preprint arXiv:2606.17833},
  year={2026}
}

@article{huang2026omg,
  title={OMG: Omni-Modal Motion Generation for Generalist Humanoid Control},
  author={Huang, Siqiao and Lee, Kun-Ying and Qiao, Dongming and He, Guanqi and Wang, Zhenyu and Li, Yitang and Zhu, Shaoting and Zhao, Hang},
  journal={arXiv preprint arXiv:2606.10340},
  year={2026}
}

@article{zhu2026hiking,
  title={Hiking in the wild: A scalable perceptive parkour framework for humanoids},
  author={Zhu, Shaoting and Zhuang, Ziwen and Zhao, Mengjie and Lee, Kun-Ying and Zhao, Hang},
  journal={arXiv preprint arXiv:2601.07718},
  year={2026}
}

@article{qin2026comprehensive,
  title={A comprehensive review of quadruped robots: Vision perception, motion control, applications and challenges},
  author={Qin, Chuan and Ruwanpathirana, Gihan and Thilakarathna, Sadeep and Wen, Hongtao and Ji, Yuxiang and Yue, Junrong and Baduge, Shanaka},
  journal={Journal of Automation and Intelligence},
  year={2026},
  publisher={Elsevier}
}
% \end{thebibliography}

\end{document}